\documentclass[runningheads]{llncs}
\usepackage[T1]{fontenc}
\usepackage{graphicx}

\usepackage{amsmath}
\usepackage{amsfonts}
\usepackage{booktabs}
\usepackage{multirow}
\begin{document}
\title{Semantics-Preserved Distortion for Personal Privacy Protection in Information Management}
%
%

\author{Jiajia Li \inst{1,2}\and
Lu Yang \inst{3} \and
Letian Peng \inst{4} \and
Shitou Zhang \inst{3} \and
Ping Wang \inst{1,2,}\thanks{$\ $  Corresponding author. This work was supported by the National Natural Science Foundation of China (No. 72074171 and No. 72374161).} \and
\\Zuchao Li \inst{3} \and
Hai Zhao \inst{5} }
%
\institute{School of Information Management, Wuhan University, Wuhan, China 
\email{wangping@whu.edu.cn}
\and
Key Laboratory of Archival Intelligent Development and Service, NAAC \and
School of Computer Science, Wuhan University, Wuhan, China 
\and
Computer Science and Engineering, University of California, San Diego \and
Shanghai Jiao Tong University
}

\maketitle              
\begin{abstract}
In recent years, machine learning—particularly deep learning—has significantly impacted the field of information management.
While several strategies have been proposed to restrict models from learning and memorizing sensitive information from raw texts, this paper suggests a more linguistically-grounded approach to distort texts while maintaining semantic integrity.
To this end, we leverage Neighboring Distribution Divergence, a novel metric to assess the preservation of semantic meaning during distortion. Building on this metric, we present two distinct frameworks for semantic-preserving distortion: a generative approach and a substitutive approach. Our evaluations across various tasks, including named entity recognition, constituency parsing, and machine reading comprehension, affirm the plausibility and efficacy of our distortion technique in personal privacy protection.
We also test our method against attribute attacks in three privacy-focused assignments within the NLP domain, and the findings underscore the simplicity and efficacy of our data-based improvement approach over structural improvement approaches.
Moreover, we explore privacy protection in a specific medical information management scenario, showing our method effectively limits sensitive data memorization, underscoring its practicality.

\end{abstract}
\section{Introduction}
The proliferation of mobile apps integrated with AI has made intelligent services widely available, ranging from social media recommendations and e-commerce to information search engines and location-based offerings.
	Online service providers use privacy agreements and certifications to collect user data, enabling personalized services that are crucial for user engagement and growth, a key competitive strategy~\cite{DBLP:journals/jeim/TangAS21}. However, such data collection can endanger user privacy, with extensive personal information gathering raising concerns about privacy violations and data breaches. This concern can affect user satisfaction and their continued use of technology.
 Online service providers thus face the challenge of alleviating users’ privacy concerns to maintain a competitive edge.
	
	Many studies on privacy-focused algorithms have been carried out~\cite{DBLP:conf/amia/DreiseitlVO01}. Current privacy-preserving strategies mostly focus on the process of training the model, such as employing differential privacy. 
 Traditional deep learning, however, requires collecting extensive user data for local model training, conflicting with privacy preservation due to the risk of information retention by these models.
 
 	In this paper, we introduce a novel privacy protection algorithm distinct from previous studies. This algorithm leverages the inherent and fundamental property of text data: semantics. Diverging from prior works that focus on new model architectures or training procedures, we pivot our focus toward data distortion, which prevents the direct exposure of sensitive data to the model while ensuring data quality for training. To apply such alteration in practical implementation, a powerful editor capable of meticulously rewriting the original text while preserving the semantics is required. In this work, we utilize Bidirectional Pre-trained Language Models (PLMs) ~\cite{DBLP:conf/naacl/DevlinCLT19} trained on Masked Language Modeling (MLM) tasks. A straightforward approach can be adopted, which involves masking sensitive details that should not be exposed to the model and using PLMs to reconstruct the masked segments. However, our observations indicate that PLMs' direct generation often lacks semantic fidelity to the original text, highlighting the necessity for a more nuanced reconstruction method.
  
  	In addressing the above challenge of preserving semantics during data distortion, we utilize Neighboring Distribution Divergence (NDD)~\cite{NDD}, a metric to evaluate semantic integrity during text edits. The NDD metric demonstrates an understanding of both syntax and semantics, enabling it to detect precise semantic differences, such as synonyms and antonyms. Leveraging the capabilities of NDD, we have implemented two frameworks, namely Generative Distortion and Substitutive Distortion, for achieving distortion while preserving the underlying semantics of the text.
   
	To explore the performance-preserving ability of our approach for natural language processing (NLP) models trained with desensitized data, we conduct experiments across three tasks: named entity recognition (NER), constituency parsing, and machine reading comprehension (MRC). Beyond these NLP tasks, we test our method against Attribute Inference Attacks (AIA) \cite{DBLP:journals/corr/abs-2210-11735}, affirming its defense capabilities against information breaches. Furthermore, we perform a detailed evaluation analysis in a medical information management scenario, and the results show our approach's effectiveness in shielding sensitive information, such as patient names and symptoms, confirming its utility in real-world applications.
 	Results from our extensive experiments validate the efficacy of our approaches, with NDD-based edits producing higher-quality texts and better performance preservation. Our main contributions can be summarized as follows: 
  	\begin{itemize}
		\item We propose semantics-preserved distortion, a novel approach for personal privacy protection within natural language processing tasks.
		\item We devise generative and substitutive frameworks for semantics-preserved distortion to robustly guard against information inference attacks.
		\item The practical efficiency of our methods is demonstrated through extensive experiments, including evaluations on NLP tasks, AIA
		tasks, and specific medical information management scenarios.
	\end{itemize}

\section{Related Work}
	To effectively deal with personal information security and privacy threats in online services, many researchers have directed their attention to the technical solutions to protect privacy. Privacy protection solutions can be divided into four categories: anonymization, encryption, differential privacy, and data modification.
 	\paragraph{Anonymization} Anonymization is a widely used method in privacy protection. It detaches the link between individual users and their rating profile~\cite{DBLP:journals/el/WuZXZXC18}, ensuring that data cannot be traced back to an individual~\cite{DBLP:journals/dke/NergizC07}. However, this method compromises data quality and system accuracy while protecting privacy. To balance privacy protection and data integrity, a new top-down thinning k-anonymous heuristic algorithm, HCE-TDR, was proposed under social network environment~\cite{DBLP:journals/lht/LiH22}. 

 	\paragraph{Encryption} 
  A swift S-Box-based encryption method for data transfer between mobile devices and providers was developed by \cite{DBLP:journals/mta/BegAAAKBK20}, with its security confirmed by histogram analysis and MLC results. To address the problem of forward secrecy, \cite{DBLP:journals/joeuc/ShuaiYWXL21} proposed a three-factor anonymous authentication scheme using a one-way hash chain.

   	\paragraph{Differential Privacy}
    Differential privacy conceals personal details while highlighting key information about user habits and behaviors, making it suitable for recommendation systems with a distributed architecture~\cite{DBLP:journals/jamia/BonomiJO20}. Building on this, a local differential privacy (LDP) component was integrated into a dynamic short-term recommendation model for small datasets to secure data privacy~\cite{DBLP:journals/wicomm/LiYYC21}. 
	
	\paragraph{Data Modification} 
 As for data modification approaches, obfuscation, data perturbation, data hiding, and randomization are commonly used. For instance, to protect the privacy of digital library users, \cite{DBLP:journals/jasis/WuLZGJS20} constructed a set of reasonable fake queries for each user query, and obscured the sensitive topic behind the user query through feature similarity. 

 	Most of the above solutions only address data privacy issues from a statistical perspective based on data distribution. Their parameters are difficult for users to interpret and understand. Also, due to their weak interpretability and semantic-preserving ability, they can hardly align with the current legal requirements, imposing limitations in both practical feasibility and the utility of the semantics of the protected data~\cite{DBLP:journals/oir/BatetS18}. 
  Protecting privacy from a semantic perspective can provide a more general, intuitive, powerful, and practical solution. However, so far, rare research on privacy protection considering data semantics has been conducted. Therefore, in this manuscript, our emphasis is on the semantic property of texts and presents a more linguistically-grounded framework to distort data while preserving core semantics.

	\section{Neighboring Distribution Divergence}
	
	\subsection{Background}
	
	In this paper, we employ two widely recognized semantic similarity metrics: perplexity and cosine similarity. 
	
	\paragraph{Perplexity}
	Perplexity is a standard metric for evaluating language models.
 It quantifies the average logarithm of the probability for each individual word in a sentence $X$ consisting of $n$ words. When assessing an MLM-based PLM, this probability is computed based on the prediction distribution for the masked positions. Let ${X}'$ be the modified sentence formed by removing the $i$-th word from the original sentence $X$ and replacing it with a placeholder [M]. This can be represented as:
 \begin{equation}
     {X}' = \left [ x_1, x_2, \cdots , x_{i-1}, [M], x_{i+1}, \cdots, x_n \right ]  
 \end{equation}
 The $\mathrm{PLM}$ predicts the distribution for the masked word, and applies the softmax function generates the probability distribution $Q$:
 \begin{equation}
 H = \mathrm{PLM} \left ( {X}' \right )
  \end{equation}
   \begin{equation}
     Q=\mathrm{Softmax}\left ( H_i \right ) 
 \end{equation}
Then, the perplexity is computed as:
\begin{equation}
    PPL = \frac{1}{n} \sum_{i=0}^{n} -log\left ( Q_{j} \right ) 
\end{equation}
where $Q_{j}$ is the probability of the $j$-th word at the $i$-th position.

High perplexity values are indicative of improbable words or unusual sentence structures, indirectly capturing semantic information. Consequently, perplexity is a widely adopted metric for evaluating the coherence and plausibility of text, enabling the detection of potential semantic anomalies within sentences.
	


	\paragraph{Cosine Similarity} 
	
	Cosine similarity is widely used to measure the semantic similarity across words, sentences, and documents. It evaluates the cosine of the angle between two vectors in a multi-dimensional space.
	
For a pair of sentences $X_{x}$ and $X_{y}$, a pre-trained encoder 
is used to encode their contextual representations into $R_{x}$ and $R_{y}$. In our experiments, we utilize a PLM-based encoder $\varepsilon$, and following the optimal representation approach in~\cite{DBLP:conf/emnlp/GaoYC21}, we employ the [CLS] token as the sentence representation. The cosine similarity between $X_{x}$ and $X_{y}$ is calculated as follows:
\begin{equation}
    \varepsilon  \left ( X_x,X_y \right ) = \frac{R_{x,CLS}\cdot R_{y,CLS} }{\left \| R_{x,CLS} \right \| \times \left \| R_{y,CLS}  \right \| } 
\end{equation}
where $R_{x,CLS}$ and $R_{y,CLS}$ are the [CLS] token representations of $R_x$ and $R_y$, respectively.

	\begin{figure*}
		\centering
		\includegraphics[width=1.0\textwidth]{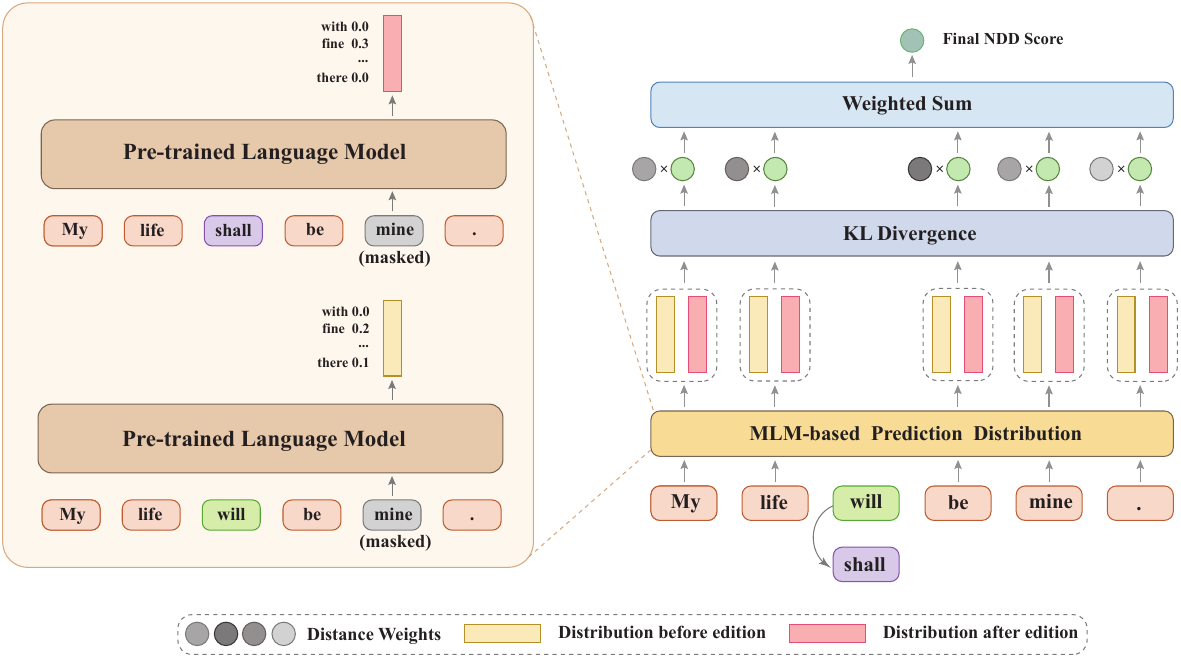}
		\caption{Computation method for Neighboring Distribution Divergence.}
		\label{fig:ndd}
	\end{figure*}
	
	\subsection{NDD Calculation}

	In this section, we expound on the computation of NDD, as detailed by \cite{DBLP:journals/corr/abs-2110-01176}. NDD quantifies the semantic disturbance on unchanged words resulting from an edit. In this context, 
 the concept of \textbf{distribution} pertains to the probability densities predicted for each word. \textbf{Divergence} represents the quantification of dissimilarity between the predicted distributions of two distinct sentences. 
 \textbf{Neighboring} represents the emphasis on words that are close to the edited spans. The procedure for calculating NDD is illustrated in Figure~\ref{fig:ndd}.

 Consider a sentence comprising $n$ words denoted as $X = \left [ x_1,x_2,\cdots, x_n \right ] $. A modification operation $E$,
 which for simplicity we assume to be a single replacement, modifies $X$. Operation $E$ substitutes the span $[x_i, \ldots, x_j]$ with $V = [v_1, \ldots, v_k]$, forming the edited sentence:
 \begin{equation}
 \check{X}  = [x_1, \ldots, x_{i-1}, v_1, \ldots, v_k, x_{j+1}, \ldots, x_n]
 \end{equation}
 
 The KL divergence is computed for the predicted distributions of the unedited adjacent words $[x_1, \ldots, x_{i-1}, x_{j+1}, \ldots, x_n]$ using MLM predictions as shown in Figure~\ref{fig:ndd}. To determine the masked distribution at the $i$-th position of a sentence $X$, we employ the following prediction method:
	

\begin{equation}
X' = [x_1, x_2, \cdots, x_{i-1}, [M], x_{i+1}, \cdots, x_n]
\end{equation}

\begin{equation}
R = \mathrm{PLM}(X')
\end{equation}
 \begin{equation}
D = Softmax(R_i)\in \mathbb{R}^{c}
\end{equation}

	We mask the $i$-th word and utilize PLM to make predictions. A softmax function generates the probability distribution $D = [d_1, d_2, \cdots, d_n]$, with $d_j$ representing the probability of the $j$-th word from the $c$-word dictionary occurring at position $i$. This prediction process is encapsulated by the function $\textrm{MLM}(\cdot)$, such that $\textrm{MLM}(X, i) = D$.
	Regarding the edit $E$, $\textrm{MLM}(\cdot)$ predicts the following distribution for the adjacent words in sentence $X$:
 \begin{equation}
     D = [d_1, \ldots, d_{i-1}, d_{j+1}, \ldots, d_n]
 \end{equation}

 We then obtain a corresponding distribution $D'$ for these words in the edited sentence $\check{X}$. The KL divergence is utilized to quantify the divergence between $D$ and $D'$:
	
	\begin{equation}
		\centering
			l = D_{KL}(d'||d) = \sum_{i=1}^c d'_i\log(\frac{d'_i}{d_i})
	\end{equation}
	
 In this context, we utilize the observed distribution $D$ from the unedited sentence and the approximate distribution $D'$ from the edited sentence. By calculating the divergence between each pair of distributions in $D$ and $D'$, we can obtain a measure of dissimilarity. To compute the final NDD, we apply a weighted sum to these divergences.
	
	\begin{equation}
		\centering
			\textrm{NDD}(X, \check{X}) = \sum_{k \in [1, \cdots, i-1, j+1, \cdots, n]} a_k l_k
	\end{equation}
where $a_k$ represents the weight assigned to each distance, defined as $\mu^{\min(|k-i|, |k-j|)}$ with $\mu \leq 1.0$. This weighting scales the divergence so that words nearer to the edit spans are given more prominence. In subsequent experiments, the weighting will be tailored to the task, typically assigning greater weight to words closer to the edits.
	
	\section{Generative and Substitutive Distortion}
	
	We detail our distortion method in this section. For a sentence $X$ with a span $X_{i:j} = [X_i, X_{i+1}, \ldots, X_j]$ containing sensitive data, we first replace the span with the mask token from the PLM.
	
	\begin{equation}
		\centering
			X_{masked} = [x_1, \cdots, x_{i-1}, \textrm{[M]}, \cdots, \textrm{[M]}, x_{j+1}, \cdots, x_n]
	\end{equation}
	
	We subsequently seek a span $S$ to occupy the masked positions with minimal semantic disturbance. In generative settings, the PLM predicts the masked positions sequentially, from which we sample a token. Conversely, in substitution settings, the center model employs pre-collected phrases for filling the masked slots.
	
	\begin{equation}
		\centering
			\check{X} = [x_1, \cdots, x_{i-1}, s_1, \cdots, s_k, x_{j+1}, \cdots, x_n]
	\end{equation}

	In both settings, we perform multiple samplings (k times) of the rewritten spans to assess the modifications in semantic meaning using the NDD metric.
	
	\begin{equation}
		\centering
			S = \mathop{\textrm{argmin}}\limits_{S}(\textrm{NDD}(X, \check{X}))
	\end{equation}
	
	\section{Experiments}
	
	
	

	
 To overcome the limited availability of textual data pertaining to privacy, we conducted our experiments to include three NLP tasks that are closely related to phrases: named entity recognition, constituency parsing, and machine reading comprehension. 
 The respective datasets utilized are the CoNLL-03 dataset \cite{DBLP:conf/CoNLL/SangM03} for named entity recognition, the Penn Treebank \cite{DBLP:journals/coling/MarcusSM94} for constituency parsing, and the SQuAD 1.0 \cite{DBLP:conf/emnlp/RajpurkarZLL16}, SQuAD 2.0 \cite{DBLP:conf/acl/RondeauH18}, and COQA \cite{DBLP:journals/tacl/ReddyCM19} datasets for machine reading comprehension. Additionally, we evaluate the Attribute Inference Attack using the Trustpilot (TP) \cite{DBLP:conf/www/HovyJS15}, AG news \cite{DBLP:conf/www/CorsoGR05}, and Blog posts (Blog) \cite{DBLP:conf/aaaiss/SchlerKAP06} datasets.
	
	
	\paragraph{Named Entity Recognition}
	
	NER aims to locate and categorize named entities within unstructured text into predefined groups like locations, person names, and organizations. 
 For NER, we experiment on the CoNLL-03 dataset.
	

	\paragraph{Constituency Parsing}
	
	Constituency parsing aims to derive a constituency-based parse tree from an unstructured sentence, delineating its syntactic structure. We utilize the Penn Treebank for our experiments and adopt a standard corpus split: sections 2-21 for training, section 22 for development, and section 23 for testing. 
	
	
	\paragraph{Machine Reading Comprehension}
	
	MRC requires models to extract answers from a given passage in response to posed questions. 

	\paragraph{AIA Defense Evaluation}
	
	To further prove the efficacy of NDD, we apply it in defending against Attribute Inference Attacks (AIA) on three datasets, namely Trustpilot, AG news, and Blog, following the preprocessing methods of \cite{DBLP:journals/corr/abs-2210-11735}. 
	
	
	\subsection{Model Setup}

	\subsubsection{Natural Language Processing Tasks}
	
	For NER and constituency parsing, We build our model following~\cite{DBLP:conf/acl/YuBP20} and~\cite{DBLP:conf/ijcai/ZhangZL20}. Instead of the conventional biaffine~\cite{DBLP:conf/iclr/DozatM17} parser, we utilize a parser called Accumulative Operation-based Induction (AOI), which has been recently proposed, to perform parsing tasks.
	Specifically, the AOI scorer has two sub-scorers, a SelfAttn scorer and a Multi-head Gathering Attention scorer (MHG). 
	SelfAttn scorer obtains dot product scores for head and dependent representations. 
 The MHG calculates global representations using multiple attention heads, concatenates it with each attention head, and chooses the max head attention scores as the attention scores for the head and depend.
 The final MHG attention scores are calculated by mutual product between head and depend scores multiplied by the sentence length. 
 The final AOI scores are the direct product of SelfAttn scores and MHG scores. 
	
	We use GloVe embeddings of $100$ dimensions~\cite{pennington-etal-2014-glove} for word representation. Additional features such as character sequences and lemmas are encoded in tensors in a $50$-dimensional space and integrated with the word embeddings. For contextualized representations, we utilize the RoBERTa-large, producing $100$-dimensional outputs that augment the input embeddings. A three-layer BiLSTM with a hidden size of $400$ processes the combined embeddings, followed by two AOI scorers with four attention heads for computing edge and label scores.
 During the training process, we minimize the Cross Entropy Loss to train the parser. We employ the Adam optimizer~\cite{DBLP:journals/corr/KingmaB14} with an initial learning rate of $10^{-3}$ to update the model's parameters. In both generative and substitutive distorting scenarios, 8 candidates are sampled for evaluation purposes. 
 
 The models are initially trained on undistorted datasets to establish performance upper bounds. Baselines are then created by training on data with spans replaced by mask tokens of the same length. Subsequently, we experiment with distortions produced by generative and substitution processes, with and without NDD control.
	
	For MRC, we test the performance of the single model on all three original datasets along with their distorted version.
	To have an equitable comparison, we adhere to the setup in \cite{DBLP:conf/naacl/DevlinCLT19} and use the BERT$_\text{large}$ as our encoder. 
	MRC models are fine-tuned for 3 epochs with a learning rate of $5e^{-5}$ and a batch size of 32.
	\begin{table}[h]
		\centering
  \caption{Performances of NER models trained on differently distorted client data.
  }
		\small
  \scalebox{0.7}{
		\begin{tabular}{ccccccc}
			\toprule
			Method & UP & UR & UF & LP & LR & LF\\
			\midrule
			Unprotected \textit{(Upper Bound)} & 93.73 & 93.62 & 93.67 & 89.08 & 88.96 & 89.02 \\
			\midrule
			Masked & 76.54 & 11.66 & 20.23 & 74.64 & 11.37 & 19.73 \\
			Generated w/o NDD & 87.29 & 72.23 & 79.05 & 76.44 & 63.25 & 69.22 \\
			Generated w/ NDD & \textbf{90.44} & \textbf{82.60} & \textbf{86.34} & \textbf{81.89} & \textbf{74.79} & \textbf{78.18} \\
			\midrule
			Substituted w/o NDD & \textbf{94.27} & 92.05 & \textbf{93.15} & 88.81 & 86.71 & 87.75 \\
			Substituted w/ NDD & 93.82 & \textbf{92.21} & 93.01 & \textbf{88.49} & \textbf{87.95} & \textbf{88.71} \\
			\bottomrule
		\end{tabular}
  }
		\label{tab:main_ner}
	\end{table}
 \begin{table}[h]
		\centering
  \caption{Evaluation of parsers with substitutive distortions at varying levels.
  }
		\small
  \scalebox{0.65}{
		\begin{tabular}{cccccccc}
			\toprule
			Method & Config. & UP & UR & UF & LP & LR & LF\\
			\midrule
			Unprotected \textit{(Upper Bound)} & - & 92.20 & 91.79 & 91.99 & 90.67 & 90.27 & 90.47 \\
			\midrule
			Substituted w/ NDD & $100\%$ P & 89.38 & 89.46 & 89.42 & 87.55 & 87.63 & 87.59 \\
			Substituted w/ NDD & $100\%$ NP & 90.61 & 90.44 & 90.52 & 88.94 & 88.78 & 88.86 \\
			Substituted w/ NDD & $33\%$ NP & \textbf{92.11} & \textbf{91.63} & \textbf{91.87} & \textbf{90.68} & \textbf{90.21} & \textbf{90.44} \\
			\bottomrule
		\end{tabular}
  }
		\label{tab:main_cp}
	\end{table}
	\subsubsection{AIA Defense Evaluation}
	
	For evaluating the proposed approach against Attribute Inference Attacks (AIA), we experiment with substituting the original dataset with the NDD metric. AIA aims to reconstruct sensitive attributes from the hidden representations generated by the model. Following \cite{DBLP:journals/corr/abs-2210-11735}, we employed five defense methods as baselines: (1) Softening prediction \cite{xu-etal-2022-student} which uses temperature parameter $\tau$ on softmax to scale probability vector, 
	(2) Prediction perturbation \cite{xu-etal-2022-student} which perturbs the original probability vector by adding Gaussian noise with a variance of $\sigma$, 
	(3) Reverse sigmoid \cite{DBLP:conf/sp/LeeEMS19} which adds random noises on the non-argmax probabilities, 
	(4) Nasty teacher \cite{DBLP:conf/iclr/MaCHYXW21} which uses Self-Undermining Knowledge Distillation to preserve accurate predictions while intentionally perturbing incorrect predictions to their maximum extent.
	and (5) Most Least \cite{DBLP:journals/corr/abs-2210-11735} which sets the predicted probabilities of the most and least likely categories to $0.5+\epsilon$ and $0.5-\epsilon$, and $0$ for others.
	To have a fair comparison, we employ the same training setting as \cite{DBLP:journals/corr/abs-2210-11735} in which we use BERT-base \cite{DBLP:conf/naacl/DevlinCLT19} as the victim and extracted models. 
 We train the models for 5 epochs using the Adam optimizer \cite{DBLP:journals/corr/KingmaB14} and set the learning rate to $2e^{-5}$.
	
	\subsection{Main Results}

	\subsubsection{Natural Language Processing Task}
	
	Table~\ref{tab:main_ner} shows the results of NER.
 Our parsers demonstrate strong performance, achieving an unlabeled F1 score of 93.67 and a labeled F1 score of 89.02. Masking alteration is found to be unreliable in preserving semantic information, as the model's performance significantly deteriorates when trained on data with masked entities. On the other hand, NDD proves beneficial in enhancing the quality of rewritten data in both generative and substitutive scenarios, resulting in improvements of 8.96 and 0.96 in the labeled F1 scores, respectively.
	
	In the constituency parsing task, we explore the substitutive scenario by replacing different phrases in the text and training parsers with this distorted data. Predominantly, we substitute noun phrases, as they are most likely to contain private information. Table~\ref{tab:main_cp} shows that, with NDD controlling, even extensive phrase substitutions do not significantly affect parser performance. Remarkably, when only a third of noun phrases are substituted, the constituency parser's performance decreases by a mere $0.03$ in the labeled F1 score.

 We select MRC for further testing. We utilized the spacy NER tool for entity identification, enabling subsequent masking/replacement operations to safeguard privacy. Results for MRC are provided in Table \ref{tab:mrc}.
	Models, trained with desensitized data, exhibit reduced performance across all datasets. Nevertheless, this performance decline is markedly less with our Substitute w/ NDD approach, suggesting it can preserve the model's generalization capability while ensuring privacy.
	
	\begin{table}[t]
		\centering
  \caption{Performances of single models on different MRC datasets.}
  \scalebox{0.7}{
		\begin{tabular}{lccccc}
			\toprule
			\multirow{2}{*}{Single Model} &  \multicolumn{2}{c}{SQuAD 1.0} & \multicolumn{2}{c}{SQUAD 2.0} & COQA \\
			\cmidrule(lr){2-3}\cmidrule(lr){4-5}\cmidrule(lr){6-6} &  EM & F$_1$ & EM & F$_1$ & F$_1$\\
			\midrule
			\multicolumn{6}{c}{\textit{Test Set}}\\
			SAN~\cite{liu2018stochastic} & 76.8 & 84.2 & 68.6 & 71.4 & - \\
			BiDAF++~\cite{DBLP:journals/corr/abs-1710-10723} & 77.6 & 84.9 & 65.6 & 68.7 & 69.5 \\
			QANet~\cite{DBLP:journals/corr/abs-1804-09541} & 80.9 & 87.8& 65.4 & 67.2 & - \\
			BERT$_\text{large}$~\cite{DBLP:conf/naacl/DevlinCLT19}& 85.1 & 91.8 & 79.9 & 83.1 & 74.4 \\
			SDNet~\cite{DBLP:journals/corr/abs-1812-03593} &  - & - & 76.7 & 79.8 & 76.6 \\
			SGNet~\cite{DBLP:journals/corr/abs-2312-05799} &  - & - & 85.1 & 87.9 & - \\
			SemBERT$_\text{large}$~\cite{DBLP:journals/corr/abs-1909-02209} & - & - & 86.1 & 88.8 & - \\
			RoBERTa$_\text{large}$~\cite{DBLP:journals/corr/abs-1907-11692} & - & - & 86.8 & 89.8 & 84.9 \\
			ALBERT$_\text{large}$~\cite{DBLP:journals/corr/abs-1909-11942} & 89.1 & 94.6 & 86.8 & 89.6 & 85.4 \\
			XLNet$_\text{large}$~\cite{DBLP:journals/corr/abs-1906-08237} & 89.9 & 95.0 & 87.9 & 90.7 & 84.6 \\
			Retr-Reader on ELECTRA~\cite{DBLP:journals/corr/abs-2001-09694} & - & - & 89.5 & 92.0 & - \\
			TR-MT (WeChatAI, 2019) & - & - & - & - & 89.3  \\
			\midrule
			\multicolumn{6}{c}{\textit{Dev Set}}\\
			Unprotected (BERT$_\text{large}$) & 90.0 & 95.2 & 88.6 & 90.9 & 88.3 \\
			\quad Substituted w/o NDD & 87.9 & 93.1 & 85.1 & 87.4 & 86.5 \\
			\quad Substituted w/ NDD & 89.6 & 95.0 & 87.8 & 90.5 & 87.7 \\
			\bottomrule
		\end{tabular}
  }
		\label{tab:mrc}
	\end{table}
	
	\subsubsection{AIA Defense Evaluation}
	\begin{table}[b]
		\centering
  \caption{Attack performance under different defenses.
  Utility means the accuracy of the victim model after adopting the defense. 
				For AIA, \textbf{higher} scores indicate better defenses. 
				All experiments are conducted on datasets with 1x queries.}
		\scalebox{0.6}{
			
				\begin{tabular}{llcccccc}
					\toprule
					\multirow{2}{*}{} & &\multicolumn{2}{c}{AG News} & \multicolumn{2}{c}{BLOG}& \multicolumn{2}{c}{TP-US}\\
					\cmidrule(lr){3-4}  \cmidrule(lr){5-6}  \cmidrule(lr){7-8} 
					& & Utility $\uparrow$ &AIA $\uparrow$ & Utility $\uparrow$ & AIA $\uparrow$ &  Utility $\uparrow$ &AIA $\uparrow$\\
					\toprule
					\multicolumn{2}{c}{No Defense} &79.99 & 15.76 &\textbf{97.07} & 34.34 & 85.53 & 36.92 \\
					\midrule
					\multicolumn{2}{l}{Softening Predictions~\cite{xu-etal-2022-student}} & 79.99  & 20.78 &97.07 &  34.91& 85.53 & \textbf{37.69} \\
					\multicolumn{2}{l}{Prediction Perturbation~\cite{xu-etal-2022-student}} &\textbf{80.03}  & 14.46  & 96.17 & 34.75 & \textbf{85.83} & 37.43\\
					\multicolumn{2}{l}{Reverse Sigmoid~\cite{DBLP:conf/sp/LeeEMS19}} & 79.99 	& 12.17 & 97.07  & 33.09 & 85.53 & 32.81 \\
					\multicolumn{2}{l}{NASTY~\cite{DBLP:conf/iclr/MaCHYXW21}} & 79.90 & 17.00	& 96.05 &	34.24	&  85.15 & 36.77\\
					\multicolumn{2}{l}{MostLeast~\cite{DBLP:journals/corr/abs-2210-11735}} & 79.99 &  17.86	&  97.07 & 34.44 & 85.53 & 37.60\\
					\midrule
					\multicolumn{2}{l}{\bf Substituted w/ NDD} & 79.70 & 15.95 & 96.84 & 34.47 & 84.92 & 37.41 \\
					\bottomrule
				\end{tabular}%
			}
			\label{tab:defense}
		\end{table}
	
	Previous experiments center on protecting privacy in specific applications/systems. We then assess actual user information leakage using three evaluation datasets: AG News, BLOG, and TP-US. We compare our proposed method with existing methods, and the detailed results are presented in Table \ref{tab:defense}. Notably, our approach, which introduces confusion to training data, is orthogonal to compared methods that modify model structures; hence, it can be combined with these to further improve efficacy.
	
	When comparing the Utility scores, the model structure-based improvement approach sporadically enhances system performance but not consistently. Our data confusion-based method has a slightly negative impact on system performance. In terms of AIA attack metrics, our approach consistently enhances defense performance compared to the ``No Defense'' approach, while other structure improvement-based approaches are not always beneficial. 

		\begin{table}
			\centering
   \caption{Evaluation of entity preservation across different generative distortion techniques.}
			\small
			\scalebox{0.7}{\begin{tabular}{ccccccc}
				\toprule
				Data & UP & UR & UF & LP & LR & LF\\
				\midrule
				Masked & 67.58 & 15.60 & 25.35 & 62.34 & 14.39 & 23.38 \\
				Generated w/o NDD & 93.06 & 86.01 & 89.40 & 92.29 & 85.30 & 88.66 \\
				Generated w/ NDD & \textbf{95.67} & \textbf{94.78} & \textbf{95.22} & \textbf{95.36} & \textbf{94.47} & \textbf{94.91} \\
				\bottomrule
			\end{tabular}}
			\label{tab:psv}
		\end{table}
		
		\subsection{Further Analysis}
		
	
		
		\begin{table}[b]
			\centering
   \caption{Comparison of the distorted results of variant configurations.}
			\small
   \scalebox{0.65}{
			\begin{tabular}{cccccccc}
				\toprule
				PLM & N-Candidate $(k)$ & UP & UR & UF & LP & LR & LF\\
				\midrule
				BERT-base-cased & 2 & 94.14 & 92.30 & 93.21 & 88.97 & 87.21 & 88.08 \\
				BERT-base-cased & 4 & 94.03 & 92.35 & 93.18 & 89.19 & 87.67 & 88.42 \\
				BERT-base-cased & 8 & 93.82 & 92.21 & 93.01 & 89.49 & 87.95 & 88.71 \\
				BERT-base-uncased & 8 & 93.65 & 92.32 & 92.98 & 88.79 & 87.67 & 88.23 \\
				BERT-large-cased & 8 & \textbf{94.56} & \textbf{92.38} & \textbf{93.45} & \textbf{89.70} & \textbf{88.17} & \textbf{88.93} \\
				\bottomrule
			\end{tabular}
			}
			\label{tab:var}
		\end{table}

		We conduct a comparative analysis of various settings to further evaluate our framework. Our primary focus was on the retention of original information during distortion. We trained our NER model using the distorted data, employing the original data as the testing set. Results, displayed in Table~\ref{tab:psv}, indicate that text masking can hardly preserve entity information. In contrast, compared to conventional generation methods not controlled by NDD, our approach significantly enhances the preservation of entity information.


		Table~\ref{tab:var} shows the performances of various configurations within our distortion framework. Generally, an increased number of selection candidates correlates with enhanced semantic preservation throughout distortion. 
  In terms of distortion, there is a trade-off between efficiency and efficacy of preservation. When calculating NDD, larger BERT models demonstrate superior performance compared to smaller ones, and cased BERT models outperform their uncased counterparts.
		
		

		
				\begin{table}
			\small
			\centering
   \caption{
   Results obtained from the {\bf Deidentified} and {\bf SPL-Deidentified} models on the generated texts.
   "w/ Name": frequency of a name token.
   "FirstName" \& "LastName": proportion of unique names generated that match names present in the MIMIC dataset.
   "EM": percentage of sentences containing a patient's name that also include one of their true (MedCAT) conditions.
   }
   \scalebox{0.7}{
			\begin{tabular}{lcccccc}
				\toprule
				Model & BioNER & w/ Name & FirstName & LastName & A@100 & EM \\
				\midrule
				BERT$_\text{base}$ & 88.32 & 84.7\% & 2.16\% & 7.72\% & 34\% & 12.17\%\\
				\midrule
				Deidentified BERT$_\text{base}$ & 90.60 & 47.9\% & 0.94\% & 3.14\% & 16\% & 23.53\% \\
				\quad \bf Substituted w/ NDD & \textbf{90.95} & 52.3\% & 1.37\% & 5.21\% & 5\% & \textbf{4.32\%} \\
				\midrule
				SPL-Deidentified BERT$_\text{base}$ & 89.24 & 59.6\% & 2.65\% & 4.56\% & 84\% & 4.17\% \\
				\quad \bf Substituted w/ NDD & 90.13 & 64.3\% & 3.42\% & 6.64\% & 23\% & 0.95\% \\
				\bottomrule
			\end{tabular}
   }
			\label{table:carlini_exp}
		\end{table}
  
		\section{Application in Medical Information Management}
		
		In the previous experimental section, we evaluate the efficacy of our proposed method in preserving performance while ensuring sensitive information is masked. Advancing this evaluation, we explore its application in real-world medical information management. PLMs like BERT and GPT have brought remarkable performance improvements in medical natural language processing tasks. 
  Typically, these models are trained on medical texts, such as clinical notes, to maintain domain consistency. And because of the high training cost of such pre-trained models, it has motivated the sharing of model parameters, such as the open pre-trained model ClinicalBERT~\cite{alsentzer-etal-2019-publicly}. 
  However, such pre-trained models may inadvertently retain sensitive information, posing significant privacy concerns, especially when trained on non-deidentified data. Consequently, this raises security issues for medical information systems due to potential privacy disclosure.

		According to the research of \cite{lehman-etal-2021-bert}, three methods are commonly used for extracting sensitive data from pre-trained models: \textbf{Prompt}, \textbf{Probe}, and \textbf{Generate}.
  Their research suggests that \textbf{Prompt} and \textbf{Probe} are less effective for this purpose, hence this study focuses on exploring the \textbf{Generate} method. Following their experimental setup, we trained two BERT models: Deidentified BERT$\text{base}$ and SPL-Deidentified BERT$\text{base}$.
  The former was trained on the Medical Information Mart for Intensive Care III (MIMIC-III) English dataset, while the latter used Electronic Health Records (EHR) dataset with patient names inserted into each sentence, to simulate a scenario where sensitive data could be more easily recovered by an adversary.
		

		We evaluate two main aspects: the performance of the pre-trained model on the BioNER task, indicating its proficiency in processing medical texts, and the other is its potential for memorizing sensitive data.
  For the BioNER task, we utilize the NCBI-Disease dataset~\cite{10.1093/bioinformatics/bty869}.
  Regarding sensitive data extraction, we focus on patient names and normalized condition mentions, mapping them to their UMLS~\cite{bodenreider2004unified} CUIs and descriptions, i.e., MedCAT.

		Table \ref{table:carlini_exp} compares BioNER and sensitive information extraction across BERT$_\text{base}$ (general domain), Deidentified BERT$_\text{base}$, and SPL-Deidentified BERT$_\text{base}$ (medical domain). 
  

		
  The accuracy of name implication is reflected by the A\@100 metric, while the name-to-condition correspondence is reflected by the EM metric. 
  The higher A@100 of SPL-Deidentified BERT$_\text{base}$ compared to Deidentified BERT$_\text{base}$ implies that the name insertion can aggravate information leakage. 
  With our proposed Substituted w/ NDD approach, the exact name extraction is markedly reduced, both in Deidentified BERT$_\text{base}$ and SPL-Deidentified BERT$_\text{base}$. 
  High EM metric correlates with privacy leakage.
  The EM metrics significantly decrease when using our Substituted w/ NDD method. 

		\section{Conclusion}
		
  In this work, we introduce a novel approach called Semantics-Preserved Distortion, which aims to defend individual privacy in natural language text. Our framework is built upon the Neighboring Distribution Divergence (NDD) metric, which accurately measures the semantic changes induced by edits. By leveraging generation and substitution techniques, we edit the sentences and select candidates that minimize NDD to ensure semantic preservation. We evaluate the effectiveness of our approach through experiments on named entity recognition, constituency parsing, and machine reading comprehension tasks. The results demonstrate that the proposed method can effectively maintain model performance while ensuring privacy. Our method can also improve defense in the evaluation against attack inference attacks. Moreover, our data-based approach can be integrated with existing model structure-based defense methods to further improve defense effectiveness. In a practical medical information management scenario, our approach has been shown to diminish the model's precise recall of patient identifiers and medical conditions.

\bibliographystyle{splncs04}
\bibliography{references}
%





\end{document}